\documentclass[nohyperref]{article}

\usepackage{microtype}
\usepackage{graphicx}
\usepackage{subfigure}
\usepackage{booktabs} 

\usepackage{hyperref}

\usepackage[accepted]{icml2022}

\usepackage{amsmath}
\usepackage{amssymb}
\usepackage{mathtools}
\usepackage{amsthm}

\usepackage[capitalize,noabbrev]{cleveref}

\theoremstyle{plain}

\theoremstyle{definition}

\theoremstyle{remark}

\usepackage[textsize=tiny]{todonotes}

\icmltitlerunning{Overcoming Adversarial Attacks for Human-in-the-Loop Applications}

\begin{document}

\twocolumn[
\icmltitle{Overcoming Adversarial Attacks for Human-in-the-Loop Applications}




\begin{icmlauthorlist}
\icmlauthor{Ryan McCoppin}{cae}
\icmlauthor{Marla Kennedy}{leidos}
\icmlauthor{Platon Lukyanenko}{cae}
\icmlauthor{Sean Kennedy}{rhw}
\end{icmlauthorlist}

\icmlaffiliation{cae}{CAE Inc. at Air Force Research Laboratory, Wright-Patterson AFB, USA}
\icmlaffiliation{leidos}{Leidos Inc. at Air Force Research Laboratory, Wright-Patterson AFB, USA}
\icmlaffiliation{rhw}{Warfighter Interactions \& Readiness Division, Air Force Research Laboratory, Wright-Patterson AFB, USA}

\icmlcorrespondingauthor{Ryan McCoppin}{ryan.mccoppin.1.ctr@afrl.af.mil}
\icmlcorrespondingauthor{Sean Kennedy}{sean.kennedy.10@afrl.af.mil}

\icmlkeywords{Machine Learning, ICML, Adversarial Robustness, Human in the loop, active learning, manipulated explanations, human vision, salience}

\vskip 0.3in
]




\begin{abstract}
Including human analysis has the potential to positively affect the robustness of Deep Neural Networks and is relatively unexplored in the Adversarial Machine Learning literature. Neural network visual explanation maps have been shown to be prone to adversarial attacks. Further research is needed in order to select robust visualizations of explanations for the image analyst to evaluate a given model. These factors greatly impact Human-In-The-Loop (HITL) evaluation tools due to their reliance on adversarial images, including explanation maps and measurements of robustness. We believe models of human visual attention may improve interpretability and robustness of human-machine imagery analysis systems. Our challenge remains, how can HITL evaluation be robust in this adversarial landscape?
\end{abstract}

\section{Introduction}
\label{intro}
The adversarial advent has greatly impacted the machine learning landscape. Adversarial images can degrade a model’s ability to perform its task. In addition to degrading ML model performance, these perturbations are often crafted to evade detection by image analysts. Additional data can be included in image analysis applications to aid the analyst; including robustness metrics and explanation maps. Even so, adversarial attacks have managed to corrupt or evade many of these additional tools. HITL defenses are needed in order to thwart attacks targeting the human element of computer vision. We present the need for further research in HITL applications and human observation in the adversarial domain.

\section{Related Work}
\label{background}

\subsection{Attacking Active Learning}
Humans are involved throughout the learning pipeline, including in curating training data. Bespoke training data is unlabeled and is labeled at substantial expense. Active learning involves more efficient labeling methods and is largely separated from the field of adversarial machine learning. While several studies have looked at misleading active learning querying \cite{miller17} there is much to explore in how active learning is affected by adversarial attacks. Specifically, how can a model query and labeling be enacted in an adversarial environment without poisoning the model.

\subsection{Manipulating Model Explanations}

\begin{figure}[ht]
    \centering
    \includegraphics[width=8cm]{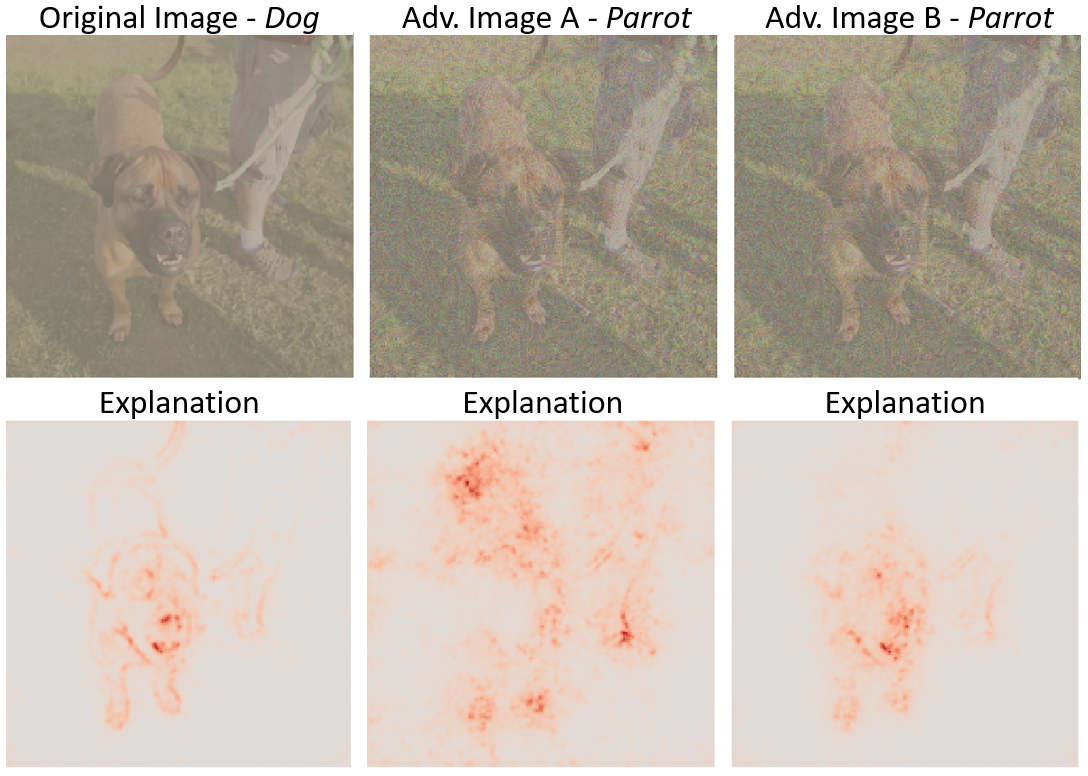}
    \caption{Adversarial images can have manipulated explanations. Image A - adversarial image; explanation reveals target class. Image B - adversarial image; manipulated explanation hides target class. Manipulation based on \cite{dombrowski19} }
  \label{fig:manip}
\end{figure}

Model utility hinges on user trust, which requires reasonable reliability measures (e.g. confidence intervals) and understanding why it made a particular decision. Past work has developed attacks that can modify classifier explanations without changing the classifier prediction. \cite{dombrowski19}. The relationship between model outputs and trust is complicated, and a recent report found that user trust following attacks is poorly studied \cite{xiongsec22}. Explanation maps which are expected to reveal adversarial images \cite{saldefense20} may be manipulated to disguise adversarial attacks and model biases. This poses obstacles not only for model trust but also for HITL evaluation (Figure \ref{fig:manip}). While solutions have been put forward which provide robustness toward manipulation \cite{dombrowski22}, the issue remains when analysts are looking at unfamiliar classes or using traditional explanatory techniques.

\section{Opportunity: Human-In-The-Loop Studies} 
\label{vision}

\subsection{Human Vision and Human Vision Models}
Because humans are not fooled by adversarial images in the same way as deep networks, humans and human vision models may contribute to adversarial robustness. Human attention models can be used to predict where humans will look when viewing a scene. Task-specific attention models can be built with gaze tracking data or crowd-sourced using interfaces that direct users to highlight (or deblur) areas of an image that are critical to their classification decision \cite{Linsley2017}. 

Disagreements between human attention models and model explanations may indicate manipulated images, low salient classes or faulty models. Taking user gaze into account could also improve model value by reducing user workload, simplifying training, or improving user performance \cite{Matzen2016}.

\subsection{Prototype Humans-in-the-Loop Tools}
\begin{figure}[ht]
    \centering
    \includegraphics[width=8cm]{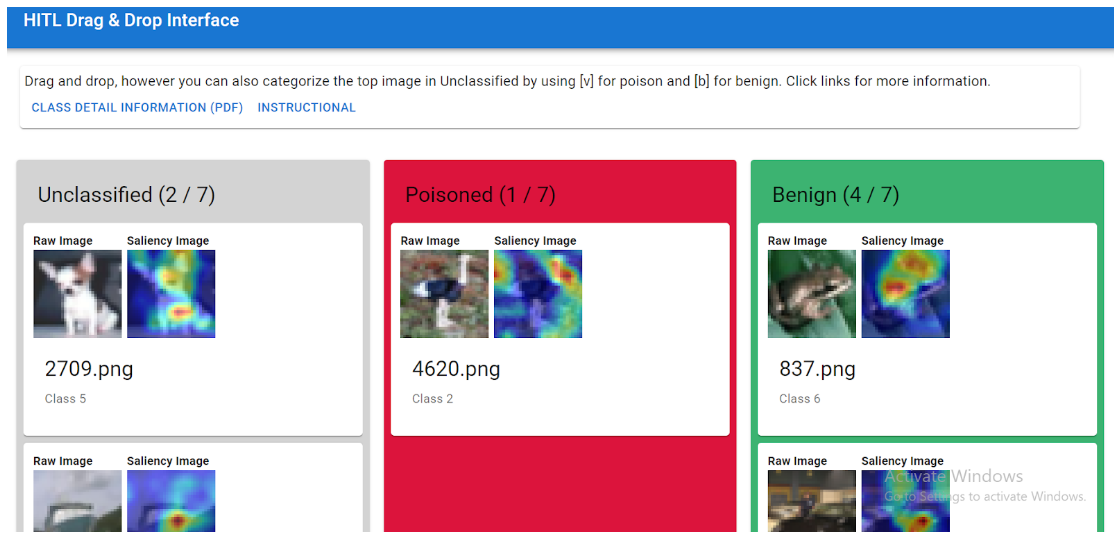}
    \caption{A HITL labeling interface with explanation maps}
  \label{fig:hitlint}
\end{figure}

Active and Interactive machine learning methods allow users to label data within an interface. This can be used to investigate attacks on active/interactive labeling, user interfaces, and be used to train models from user responses. One of our prototype HITL tools examines how users can contribute to adversarial or poisoned image detection. Users assign ‘cards’ that contain images and metadata to ‘poisoned’ or ‘benign’ categories, as seen in Figure \ref{fig:hitlint}.  Visual explanations of a classifier’s decision, obtained via Grad-CAM \cite{gradcam2017}, provide the user with regions of an image the ML model focuses on during classification. As user annotated data is collected, a poison detection ML model is updated via active learning. With this tool, we aim to explore:
\begin{itemize}
\item How do adversarial detection models compare to analyst detection capability?
\item What explanations do users find useful and convincing?
\item Do more robust explanations improve human performance?
\item Which attacks are really invisible? What tools can make them more visible?
\end{itemize}

\section{Challenges}
\label{wrapup}

There are many challenges in HITL research concerning adversarial robustness that have yet to find answers. These concerns include visualizing perturbations, which explanation maps are robust and beneficial, the reliability of active learning query methods and which additional metrics may be beneficial to the analyst. 

 Generally, users are unable to detect adversarial images and thus rely on ``explanation maps'' to provide visibility. These maps can also be adversarial distorted \cite{dombrowski19}. If explanation maps are not always reliable or truthful, what information can be provided to the user in order to detect adversarial images and improve model robustness? In addition, is it possible to gain a robustness to these distortions in a similar manner to adversarial training? We believe human vision models may provide a key in detecting and interpreting adversarial images. However, recent research has shown that deep models of human attention are also vulnerable to adversarial attack \cite{che2020}. Can we improve the adversarial robustness of these models and combine human and machine attention to identify adversarial images? Research has already begun in adversarial robustness, but not in how it relates to HITL evaluation and how analysts are able to use more robust explanations.

\section*{Acknowledgements}
The opinions expressed herein are solely those of the authors and do not necessarily represent the opinions of the United States Government, the U.S. Department of Defense, the Department of the Air Force, or any of their subsidiaries or employees.
This research was funded by a Commander's Research and Development Fund grant from the Air Force Research Laboratory.
DISTRIBUTION STATEMENT A.

\nocite{cifar09}
\nocite{russa15}

\bibliography{main_paper}

\begin{thebibliography}{11}
\providecommand{\natexlab}[1]{#1}
\providecommand{\url}[1]{\texttt{#1}}
\expandafter\ifx\csname urlstyle\endcsname\relax
  \providecommand{\doi}[1]{doi: #1}\else
  \providecommand{\doi}{doi: \begingroup \urlstyle{rm}\Url}\fi

\bibitem[Che et~al.(2020)Che, Borji, Zhai, Ling, Li, and Le~Callet]{che2020}
Che, Z., Borji, A., Zhai, G., Ling, S., Li, J., and Le~Callet, P.
\newblock A new ensemble adversarial attack powered by long-term gradient
  memories.
\newblock \emph{AAAI Conference on Artificial Intelligence}, 34:\penalty0
  3405--3413, Apr. 2020.

\bibitem[Dombrowski et~al.(2019)Dombrowski, Alber, Anders, Ackermann, Müller,
  and Kessel]{dombrowski19}
Dombrowski, A.~K., Alber, M., Anders, C., Ackermann, M., Müller, K.~R., and
  Kessel, P.
\newblock Explanations can be manipulated and geometry is to blame.
\newblock In \emph{Advances in Neural Information Processing Systems 32}.
  Elsevier, 2019.

\bibitem[Dombrowski et~al.(2022)Dombrowski, Anders, Müller, and
  Kessel]{dombrowski22}
Dombrowski, A.~K., Anders, C.~J., Müller, K.~R., and Kessel, P.
\newblock Towards robust explanations for deep neural networks.
\newblock In \emph{Pattern Recognition}. Elsevier, 2022.

\bibitem[Krizhevsky \& Hinton(2009)Krizhevsky and Hinton]{cifar09}
Krizhevsky, A. and Hinton, G.
\newblock Learning multiple layers of features from tiny images.
\newblock Technical report, Computer Science Department, University of Toronto,
  Toronto, Canada, 2009.

\bibitem[Linsley et~al.(2017)Linsley, Eberhardt, Sharma, Gupta, and
  Serre]{Linsley2017}
Linsley, D., Eberhardt, S., Sharma, T., Gupta, P., and Serre, T.
\newblock What are the visual features underlying human versus machine vision?
\newblock In \emph{2017 IEEE International Conference on Computer Vision
  Workshops (ICCVW)}, pp.\  2706--2714, 2017.

\bibitem[Matzen et~al.(2016)Matzen, Haass, Tran, and McNamara]{Matzen2016}
Matzen, L., Haass, M., Tran, J., and McNamara, L.
\newblock Using eye tracking metrics and visual saliency maps to assess image
  utility.
\newblock \emph{Electronic Imaging}, 2016:\penalty0 1--8, 02 2016.

\bibitem[Miller et~al.(2017)Miller, Hu, Qiu, and Kesidis]{miller17}
Miller, D.~J., Hu, X., Qiu, Z., and Kesidis, G.
\newblock Adversarial learning: A critical review and active learning study.
\newblock In \emph{27th International Workshop on Machine Learning for Signal
  Processing}, pp.\  1--6. IEEE, 2017.

\bibitem[Russakovsky et~al.(2015)Russakovsky, Deng, Su, Krause, Satheesh, Ma,
  Huang, Karpathy, Khosla, Bernstein, Berg, and Fei-Fei]{russa15}
Russakovsky, O., Deng, J., Su, H., Krause, J., Satheesh, S., Ma, S., Huang, Z.,
  Karpathy, A., Khosla, A., Bernstein, M., Berg, A.~C., and Fei-Fei, L.
\newblock Imagenet large scale visual recognition challenge.
\newblock In \emph{International Journal of Computer Vision (IJCV)}. Springer,
  2015.

\bibitem[Selvaraju et~al.(2017)Selvaraju, Cogswell, Das, Vedantam, Parikh, and
  Batra]{gradcam2017}
Selvaraju, R.~R., Cogswell, M., Das, A., Vedantam, R., Parikh, D., and Batra,
  D.
\newblock Grad-cam: Visual explanations from deep networks via gradient-based
  localization.
\newblock In \emph{2017 IEEE International Conference on Computer Vision
  (ICCV)}, pp.\  618--626, 2017.

\bibitem[Xiong et~al.(2022)Xiong, Buffett, Iqbal, Lamontagne, Mamun, and
  Molyneaux]{xiongsec22}
Xiong, P., Buffett, S., Iqbal, S., Lamontagne, P., Mamun, M., and Molyneaux, H.
\newblock Towards a robust and trustworthy machine learning system development:
  An engineering perspective.
\newblock In \emph{Journal of Information Security and Applications}, pp.\ ~65.
  Elsevier, 2022.

\bibitem[Ye et~al.(2020)Ye, Chen, Liu, Wang, and Jiang]{saldefense20}
Ye, D., Chen, C., Liu, C., Wang, H., and Jiang, S.
\newblock Detection defense against adversarial attacks with saliency map.
\newblock In \emph{International Journal of Intelligent Systems}. Wiley Online
  Library, 2020.

\end{thebibliography}
\bibliographystyle{icml2022}

\end{document}